\documentclass{article}

\usepackage{graphicx}
\usepackage{subcaption}
\usepackage{booktabs} 

\usepackage{hyperref}

\usepackage[accepted]{icml2026}

\usepackage{amsmath}
\usepackage{amssymb}
\usepackage{mathtools}
\usepackage{amsthm}

\usepackage[capitalize,noabbrev]{cleveref}

\theoremstyle{plain}

\theoremstyle{definition}

\theoremstyle{remark}

\usepackage[textsize=tiny]{todonotes}

\icmltitlerunning{ICML Workshop on Pluralistic Alignment: Socially Grounded Agentic AI}

\begin{document}

\twocolumn[
  \icmltitle{Socially Grounded Agentic AI:\\ Coordinating Plural Perspectives through Social Theory}



  \icmlsetsymbol{equal}{*}

  \begin{icmlauthorlist}
    \icmlauthor{Matt Ratto}{equal,a,b4,b5}
    \icmlauthor{Abhishek Moturu}{equal,b4,b1,b2,b3}
    \icmlauthor{Daniel Silver}{equal,c}
  \end{icmlauthorlist}

  \icmlaffiliation{a}{Faculty of Information, University of Toronto, Ontario, Canada}
  \icmlaffiliation{b1}{Department of Computer Science, University of Toronto, Ontario, Canada}
  \icmlaffiliation{b2}{Vector Institute for Artificial Intelligence, Ontario, Canada}
  \icmlaffiliation{b3}{Temerty Centre for Artificial Intelligence Education and Research In Medicine, University of Toronto, Ontario, Canada}
  \icmlaffiliation{b4}{Massey College, Ontario, Canada}
  \icmlaffiliation{b5}{Schwartz Reisman Institute for Technology and Society, University of Toronto, Ontario, Canada}
  \icmlaffiliation{c}{Department of Sociology, University of Toronto Scarborough, Ontario, Canada}

  \icmlcorrespondingauthor{Matt Ratto}{matt.ratto@utoronto.ca}
  \icmlcorrespondingauthor{Abhishek Moturu}{moturuab@cs.toronto.edu}

  \icmlkeywords{Machine Learning, ICML}

  \vskip 0.3in
]



\printAffiliationsAndNotice{}  


\begin{abstract}
As AI systems are deployed across increasingly diverse social contexts, alignment can no longer be framed as the optimization of a single, unified set of values. Instead, systems must be able to recognize, represent, and respond to multiple legitimate perspectives. This has led to growing interest in pluralistic alignment, which seeks to move beyond one-size-fits-all models of appropriate behaviour. However, current approaches often lack a clear account of how values are socially organized, contested, and coordinated in practice. In this paper, we argue that social theory provides essential conceptual and design resources for addressing these challenges. Drawing on established traditions in sociology, we show how perspectives can be understood as structured by roles, shaped through interaction, and distributed across fields of power and expertise. We translate these insights into concrete implications for AI system design, including role-based representations, structured coordination among perspectives, and context-sensitive evaluation. For agentic systems, this requires aligning not only final outputs, but also the role activations, deliberative traces, aggregation rules, and feedback loops through which those outputs are produced. Our contribution is to reposition pluralistic alignment as a problem of socially grounded coordination rather than output diversification. We outline a design space for systems that engage multiple perspectives in structured and accountable ways, and we identify directions for future work to implement and empirically evaluate these approaches in real-world settings.
\end{abstract}

\section{Introduction}

As AI systems become more widely embedded in public, institutional, and everyday decision-making contexts, alignment can no longer be understood as the task of fitting models to a single, universal account of human preference or value. Large language models are increasingly used by people who differ in culture, political orientation, professional role, institutional position, and lived experience. Under these conditions, a system that appears ``aligned'' from one standpoint may fail, exclude, or misrepresent another. This has motivated growing interest in \textit{pluralistic alignment}: the design of AI systems that can recognize, represent, and respond to multiple legitimate perspectives rather than collapsing disagreement into a single preferred answer \citep{sorensen2024roadmap, kirk2023personalisation}.

Recent work has begun to formalize this problem. Sorensen et al. identify three prominent strategies for pluralistic alignment: \textit{Overton pluralism}, in which a model presents a bounded range of reasonable responses; \textit{steerable pluralism}, in which users or institutions can guide a model toward particular values or perspectives; and \textit{distributional pluralism}, in which model outputs reflect the distribution of views within some target population \citep{sorensen2024roadmap}. These approaches provide important technical vocabulary for moving beyond one-size-fits-all alignment. They also make clear that pluralistic alignment is not only a problem of model behaviour, but of social representation: whose values are represented, how disagreement is structured, when steering is legitimate, and what counts as an acceptable or reasonable position. This concern also connects pluralistic alignment to social choice, where the central problem is how to aggregate divergent human input into collective choices about model behaviour~\citep{conitzer2024social}. This does not imply that all expressed preferences should be represented symmetrically: pluralistic systems still require explicit safety, rights-based, and domain-specific constraints on which perspectives may be surfaced, weighted, or acted upon.

Agentic AI makes pluralistic alignment harder because systems no longer merely answer isolated prompts. They maintain state, invoke tools, coordinate subtasks, delegate to components, and act across multi-step trajectories. As a result, pluralistic alignment must govern not only what a model says, but how roles are activated, how conflicts are resolved, when tools are used, and when humans are brought back into the loop.

This paper argues that social theory offers underused resources for addressing these challenges. Sociological and social-theoretical traditions have long studied how values, roles, norms, conflicts, institutions, and publics are organized. They provide concepts for understanding not only what people believe, but how perspectives are formed, stabilized, contested, and coordinated in social life. In this sense, social theory can help pluralistic alignment move beyond the enumeration of viewpoints toward richer models of situated judgment, institutional role, deliberation, power, and interaction.

We develop this argument by examining how key social-theoretical concepts can be operationalized within current AI design practices. Mead's~\citep{mead1934mind} concept of the generalized other helps reframe Overton pluralism as a problem of modeling role-structured fields of expectation. Habermasian~\citep{habermas1985theory} theories of deliberation help reinterpret steerability as the design of discursive conditions under which claims can be justified and challenged. Bourdieu's~\citep{bourdieu1984social} theory of fields foregrounds the power relations that shape distributional pluralism, asking not only what a population believes but how that population is constructed and whose positions are made legible. 

This paper contributes a socially grounded design framework for pluralistic alignment. Socially grounded pluralistic alignment is not another aggregation rule, but an account of the social conditions under which aggregation, steering, and evaluation become legitimized. First, we reinterpret three existing pluralistic alignment strategies (Overton, steerable, and distributional pluralism) as problems of role representation, deliberative process, and field-aware aggregation. Second, we translate Mead, Habermas, and Bourdieu into an operational pipeline for agentic AI systems: role-indexed representation, structured multi-agent deliberation, provenance-sensitive aggregation, and trajectory-level audit. Third, we propose evaluation criteria for socially grounded pluralism, including role fidelity, perspective coverage, deliberative quality, provenance transparency, field sensitivity, and escalation appropriateness. Together, these contributions position pluralistic alignment as a problem of accountable social coordination rather than output diversification. Future work will implement and evaluate these mechanisms in applied domains where plural values are unavoidable, including healthcare coordination, institutional advising, and public-facing AI services.

\section{From Overton Windows to Generalized Others}

One of the most prominent approaches to pluralistic alignment is what \citet{sorensen2024roadmap} describe as \textit{Overton pluralism}: the idea that AI systems should surface a range of responses that fall within a bounded space of socially acceptable or ``reasonable'' viewpoints. This approach implicitly draws on the notion of an Overton window, a shifting range of ideas that are considered legitimate within public discourse. In practice, Overton-style pluralism is often implemented by prompting models to generate multiple perspectives, filtering outputs through safety or acceptability criteria, or fine-tuning models on curated datasets that reflect a diversity of positions \citep{kirk2023personalisation}.

While useful, this approach faces a fundamental limitation: it treats plurality as a problem of listing. That is, it assumes that the task is to identify and list a set of acceptable viewpoints. This leads to several well-known challenges. First, it is difficult to determine the boundaries of the Overton window: what counts as ``reasonable,'' and for whom? Second, enumerations are inherently unstable: as contexts shift, new viewpoints emerge while others recede. Third, list-based approaches obscure the structure that gives rise to viewpoints in the first place, reducing socially embedded perspectives to decontextualized statements.

We argue that these limitations arise because Overton pluralism lacks an explicit model of the social organization of perspective. To address this, we draw on George Herbert Mead's concept of the \textit{generalized other} \citep{mead1934mind}. For Mead, social action is not guided by isolated individual preferences but by an internalized understanding of the expectations, roles, and norms that constitute a social field. The generalized other is not a collection of viewpoints but a structured set of relations among positions (roles such as professional, institutional, or cultural identities) that shape how actors interpret situations and formulate responses.

Reframed in these terms, the problem of Overton pluralism becomes one of modeling role-structured fields of expectation. Rather than asking a model to produce multiple ``opinions,'' we ask it to generate responses grounded in specific social positions, each carrying its own constraints, obligations, and epistemic commitments. For example, in healthcare, a clinician, an administrator, and a patient advocate do not just hold different ``views''; they operate within distinct institutional logics that shape what counts as a relevant concern, a valid justification, or an acceptable outcome.

This shift from viewpoints to roles has several implications for AI system design. First, it suggests that pluralistic outputs should be indexed to roles rather than presented as anonymous alternatives. Role-structured prompting can encode institutional positions directly within system instructions, enabling models to produce responses that reflect situated reasoning rather than abstract variation. Second, it highlights the importance of relational representation: roles do not exist in isolation but in structured relation to one another (e.g., authority, accountability, conflict, collaboration). Models should therefore represent not only multiple perspectives, but also the relationships among them. Third, it opens the possibility of integrating institutional constraints into model behaviour, aligning outputs with the norms and responsibilities associated with specific positions.

It is important to note that a role is not a demographic profile, stylistic persona, or prompt-level character. It is a situated position within a field of action, defined by its obligations, sources of knowledge, scope of authority, relation to other positions, and escalation conditions. In implementation terms, a role representation should, at least, include: 
\begin{enumerate}
    \item the institutional or social position being represented,
    \item the domain in which the role is relevant,
    \item the obligations and constraints attached to that role,
    \item the sources or forms of evidence it is licensed to use,
    \item its relation to other roles, and
    \item the conditions under which it should defer, challenge, or escalate.
\end{enumerate}

This distinguishes socially grounded role representation from persona prompting. Rather than imitating a type of person, role representation requires the model to reason from a position with explicit authority, constraints, and accountability. Role selection is therefore a governance decision, not an unconstrained inference from user text. Systems should record why a role was activated, what assumptions were made, and whether that activation was contested.

This approach also avoids reifying social categories. Roles should be used minimally, contextually, and revisably, without inferring protected identities unless they are relevant and explicitly provided. Systems should distinguish institutional roles from demographic identities and update role definitions when stakeholders identify misrepresentation. Perspectives that are harmful, oppressive, or outside domain-specific constraints need not be represented as equal deliberative agents, but can be treated as risks, objections, or boundary conditions addressed through safety, rights-based, or institutional safeguards.

\subsection*{Operationalizing generalized others}
While complex, these ideas can be operationalized within existing AI pipelines. For example, role or persona metadata can be incorporated into prompts or training datasets, allowing models to condition their responses on social position and generate outputs that reflect distinct reasoning styles and linguistic structures \citep{tseng2024two, hu2024quantifying, kirk2023personalisation}. Retrieval-augmented generation (RAG) systems can be extended to include role-indexed corpora, enabling models to condition generation on contextually appropriate sources and thereby produce outputs aligned with specific social or institutional perspectives \citep{lewis2020retrieval, wu2024retrieval}. Multi-agent architectures can instantiate roles as distinct agents, enabling interaction among perspectives rather than static enumeration. Together, these approaches move Overton pluralism from a problem of filtering outputs to one of modeling the social structure that produces them.

This reframing also aligns with broader shifts in AI research toward socially grounded and context-sensitive systems. As argued in recent work on pluralistic alignment, the challenge is not simply to increase diversity in outputs, but to represent and coordinate perspectives in ways that are legible, accountable, and responsive to context \citep{sorensen2024roadmap}. By drawing on Mead's generalized other, we provide a conceptual and operational foundation for this shift, positioning Overton pluralism as the first step toward a more fully social model of alignment.

\section{Steerability as Deliberative Design}

A second major approach to pluralistic alignment is \textit{steerable pluralism}, in which users or system designers guide model outputs toward particular value frameworks \citep{sorensen2024roadmap}. In practice, this is typically implemented through prompt conditioning, system instructions, or reinforcement learning from human feedback (RLHF), enabling models to reflect specified preferences such as tone, domain norms, or political orientation \citep{kirk2023personalisation, ouyang2022training}. While effective, these approaches treat steerability primarily as a control problem: how to adjust model behaviour in response to inputs. This framing is limited in settings where values are incomplete, conflicting, or must be negotiated among stakeholders.

To address this, we draw on the Habermasian theory of communicative action, which frames coordination as the outcome of deliberation rather than unilateral specification \citep{habermas1985theory}. From this perspective, steerability becomes a problem of designing the conditions under which perspectives can be articulated, contested, and revised. This shift is increasingly reflected in recent AI systems that model reasoning as interaction. Multi-agent LLM frameworks such as AutoGen and CAMEL represent tasks as structured exchanges among role-differentiated agents \citep{wu2024autogen, li2023camel}, while debate-based alignment approaches use adversarial interaction to improve reasoning quality and factual accuracy \citep{irving2018debate, du2024improving}. Similarly, work on tool-augmented reasoning and ReAct-style architectures shows that interleaving reasoning with action and verification improves robustness in multi-step tasks \citep{yao2022react}.

Across these approaches, a common pattern emerges: steerability is achieved by structuring interaction. Agents operate with differentiated roles, exchange claims supported by evidence, critique one another, and converge through defined decision procedures. Computational argumentation research further reinforces this view, showing that structured formats such as claim-evidence-rebuttal support transparency and evaluability in AI-mediated decision-making \citep{lawrence2019argument}. 

This perspective suggests that deliberative architectures can be understood as composed of four tightly coupled elements: role-differentiated agents, interaction protocols governing exchange, structured reasoning formats, and explicit closure mechanisms such as synthesis, arbitration, or escalation. Unlike prompt-based steering, these systems produce outcomes through interaction traces, making the reasoning process inspectable and revisable.


This reframing shifts steerability along three dimensions. First, it replaces direct control with structured interaction, recognizing that alignment often requires coordination among partially conflicting values. Second, it foregrounds process legitimacy, enabling users to inspect how conclusions are reached rather than only what is produced. Third, it enables adaptive steering, in which systems negotiate responses dynamically rather than relying on fixed value specifications. These properties align with emerging work on agent governance and evaluation, which treats interaction protocols and reasoning traces as central to system reliability \citep{Anthropic2026Evals, kim2025beyond}.

\subsection*{Operationalizing Deliberative Steerability}

The shift from control to coordination can be implemented within existing agentic AI pipelines using a small set of architectural extensions. Role-differentiated agents can be instantiated through structured system prompts or separate model instances, each encoding distinct perspectives (e.g., domain expertise, stakeholder position, or institutional constraint). Additionally, interaction protocols can be enforced through constrained generation loops, where agents must produce outputs in structured formats (e.g., claim-evidence-justification) and respond to critiques before proceeding. This builds on existing patterns in multi-agent orchestration and tool-augmented reasoning \citep{wu2024autogen, yao2022react}. Also, deliberative state can be maintained using external memory or logging mechanisms that record exchanges, enabling cumulative reasoning and trajectory-level evaluation. Finally, closure mechanisms, such as arbitration agents, voting rules, or escalation triggers, can be implemented at the orchestration layer to ensure that interaction produces actionable outcomes. These components can be composed using existing frameworks for agent coordination, requiring minimal modification to underlying models.

Together, these design elements allow deliberative steerability to be realized as a layer of structured interaction and control. This makes the approach compatible with current LLM systems while enabling more robust handling of conflicting perspectives and dynamic contexts.

\section{Distributional Pluralism as Field Representation}

A third major strategy for pluralistic alignment is \textit{distributional pluralism}, which aims to align model outputs with the distribution of values or opinions within a target population \citep{sorensen2024roadmap}. Rather than presenting a bounded set of acceptable viewpoints (Overton pluralism) or enabling users to steer outputs toward particular perspectives (steerable pluralism), distributional approaches seek to reflect how beliefs are actually distributed across groups. In principle, this allows models to represent not only diversity but also the relative prevalence of different positions.

However, this approach raises a set of fundamental challenges. First, it is unclear how to define the relevant population: should distributions reflect global users, local communities, domain experts, or historically marginalized groups? Second, empirical distributions are often shaped by structural inequalities, meaning that naive replication may reproduce existing biases and exclusions. Third, distributions themselves are not static, but shift over time and across contexts, making them difficult to represent reliably. As a result, distributional pluralism risks presenting a seemingly neutral account of plurality that obscures the social processes through which that plurality is produced.

To address these challenges, we draw on Pierre Bourdieu's theory of social fields \citep{bourdieu1984social}. For Bourdieu, social life is organized into relatively autonomous fields (e.g., medicine, law, education), each structured by relations of power, forms of capital (economic, cultural, social), and struggles over legitimacy. Within a field, positions are not equivalent: some actors have greater authority to define what counts as valid knowledge or acceptable practice. Crucially, distributions of belief or preference cannot be understood independently of these underlying structures.

Reframed in these terms, distributional pluralism becomes a problem of \textit{field representation}. Rather than treating populations as flat aggregates of individuals, we model them as structured configurations of positions, each associated with different forms of authority, expertise, and marginalization. The key question shifts from ``what do people believe?'' to ``how are perspectives distributed across positions within a field, and how are those positions constituted?''

\subsection*{Operationalizing Field-Aware Alignment}

A field-based view of pluralistic alignment can be implemented through a small set of coordinated design mechanisms. First, systems can incorporate \textit{population-aware routing}, directing inputs to models or agents associated with particular social positions (e.g., professional roles, stakeholder groups, or demographic contexts). This extends mixture-of-experts architectures by introducing socially meaningful routing criteria rather than purely task-based specialization. Second, outputs can be combined through \textit{position-weighted aggregation}, in which contributions are weighted not only by empirical prevalence but also by normative considerations such as expertise, equity, or institutional role. Third, systems can expose \textit{provenance information}, making visible which populations, datasets, or perspectives contributed to a given output.

Crucially, these mechanisms must operate dynamically. Population representations should be treated as context-dependent and evolving, updated through feedback, interaction, and external data rather than fixed at training time. This makes explicit that distributional alignment is inherently normative: decisions about which populations to include, how to weight perspectives, and how to handle inequality require policy choices rather than purely statistical inference. These choices can be operationalized through techniques such as corrective weighting, counterfactual simulation of alternative distributions, or user-selectable population views.

A simple example illustrates the approach. In a policy recommendation system, a field-aware design would distinguish among stakeholders (e.g., policymakers, experts, affected communities), model their relative positions, weight their contributions based on both expertise and equity considerations, and expose these assumptions to users. The resulting output is not a single ``representative'' answer, but a structured representation of the field within which decisions are made.

Taken together, these mechanisms shift distributional alignment from representing flat populations to modeling structured social fields. They enable systems to move beyond reproducing observed distributions toward coordinating perspectives in ways that are context-sensitive, transparent, and open to revision.

\section{Interaction and Evaluation in Pluralistic Systems}

The preceding sections have reframed pluralistic alignment in terms of role-structured perspectives (Mead), deliberative processes (Habermas), and field-aware population representations (Bourdieu). Together, these approaches shift alignment from static output generation toward structured coordination among socially grounded positions. However, a remaining challenge concerns how these structures are enacted and evaluated in practice.

Pluralistic alignment is not only a property of model outputs or internal representations; it is also an emergent feature of interaction between users, models, and institutional contexts. In real-world deployments, systems must respond to ongoing exchanges, adapt to feedback, and maintain coherence across sequences of interaction. This suggests that alignment must be assessed not only at the level of individual responses, but at the level of \textit{interaction trajectories}.

Work in social theory, including Randall Collins' interaction ritual theory, emphasizes that shared understanding and legitimacy emerge through repeated, structured encounters rather than isolated statements \citep{collins2014interaction}. Related traditions in sociology and pragmatism similarly treat judgment as an ongoing, situated accomplishment rather than a one-time output. This perspective suggests a key design implication for pluralistic alignment: systems must account for how perspectives are sustained, revised, and coordinated across interaction over time, not just within single responses.

\subsection{Operational Implications}

Recent technical work on agentic systems and multi-step reasoning reinforces this shift. Multi-agent frameworks and debate-based approaches already evaluate performance over trajectories of interaction rather than isolated outputs \citep{wu2024autogen, du2024improving}. Building on this, we highlight two tightly coupled implications for system design.

\textbf{Interaction traces and trajectory-level evaluation.}  
Pluralistic alignment should treat interaction histories as primary alignment objects rather than focusing solely on final outputs. Recent work on LLM agents shows that evaluating performance requires going beyond answer correctness to assess full reasoning trajectory, including intermediate steps, tool use, and adaptation over time \citep{liu2023agentbench, kim2025beyond, he2025traject}. Emerging benchmarks and evaluation frameworks explicitly measure properties such as coherence across multi-turn interactions, intermediate decision correctness, and robustness over extended reasoning sequences \citep{he2025traject, li2026atbench}. 

This shift reflects a broader transition in evaluation practice: trajectory-level metrics capture aspects of system behaviour such as efficiency, hallucination, and adaptivity that are invisible to outcome-only evaluation \citep{kim2025beyond}. Similarly, recent work on agent evaluation emphasizes that the full execution path, including reasoning steps, tool calls, and interaction structure, must be observable to support debugging, reliability, and alignment assessment \citep{gritta2026process, zhuge2024agent}. 

Treating interaction traces as primary evidence makes pluralistic alignment easier to inspect and audit. Because these traces record how roles are activated, how claims are challenged, and how positions are revised over time, they allow stakeholders to evaluate more than the final output. They make it possible to assess how the system reached its conclusion, bringing technical evaluation closer to broader standards of transparency and procedural accountability.

\textbf{Adaptive orchestration as alignment mechanism.}  
Interaction also provides a basis for ongoing adjustment. Recent work on LLM agents and deployed ML systems emphasizes that reliable behaviour depends on continuous monitoring and feedback, rather than one-time training or evaluation \citep{shankar2021towards, Anthropic2026Evals}. Signals such as persistent disagreement between agents, repeated human overrides, or user dissatisfaction can be leveraged to modify system behaviour at the orchestration level, for example, by altering interaction protocols, reweighting perspectives, or triggering escalation to human review. This aligns with approaches to continual alignment and online adaptation, where models are updated or guided through feedback loops that incorporate new information over time \citep{leike2018scalable, bai2022constitutional}.

In agentic systems, these adjustments increasingly operate above the level of model parameters, within the control and coordination layer or ``harness'' that governs how components interact. Recent frameworks for agent orchestration and evaluation highlight the importance of feedback-driven adaptation across extended trajectories, where system performance depends on how effectively it responds to changing conditions and accumulated signals \citep{wu2024autogen, kim2025beyond}. From this perspective, alignment is not a static property fixed at training time, but a dynamic process maintained through structured interaction and feedback loops. Treating adaptation as an intrinsic feature of system operation enables AI systems to remain responsive, robust, and context-sensitive over time.

\subsection{Position within the Framework}

Importantly, this interactional perspective provides a way of \textit{integrating and evaluating} the three strategies discussed above. Role-based representations, deliberative processes, and field-aware distributions are enacted through interaction and must be assessed in those terms.

By focusing on interaction trajectories, we can better understand how plural perspectives are coordinated in practice, how legitimacy is established, and where breakdowns occur. This perspective also creates a bridge to empirical work: systems can be instrumented to collect interaction data, enabling the systematic study of pluralistic alignment in deployment settings.

More concretely, this trajectory-level view makes it possible to define evaluation criteria for each component of the framework. For role-based representations, interaction traces can reveal whether the system activated the relevant social roles, preserved their distinct obligations, and avoided collapsing them into generic personas. For deliberative processes, traces can show whether claims were supported, challenged, revised, and closed through an explicit decision rule. For field-aware distributions, traces can expose which populations or forms of expertise were invoked, how they were weighted, and whether unresolved exclusions or power asymmetries remained visible. Making this process auditable lets stakeholders inspect not only what answer a system produced, but how plural perspectives were selected, coordinated, and resolved. Thus, interaction allows socially grounded alignment to become observable, measurable, and ultimately improvable.

\section{Example: Clinical Triage as Social Coordination}
\label{clinical}

Consider a patient-facing scheduling assistant that is asked: ``I have chronic pain that has been getting worse. The earliest specialist appointment is in six weeks, but I cannot afford to take additional time off work. Should I wait for the appointment or try to get seen sooner?''

This request does not have a single objectively correct answer. It involves potentially competing considerations related to clinical risk, patient burden, institutional resource allocation, and access to care. A baseline pluralistic system might respond by generating several alternative answers: one emphasizing medical caution, another emphasizing work and financial constraints, and another emphasizing the difficulty of obtaining earlier appointments. While these responses are diverse, they do not explain how competing considerations should be weighed, which perspectives are legitimate decision participants, or how conflicts between them should be resolved.

A socially grounded agentic system instead treats the request as a problem of social coordination.

\begin{enumerate}
\item \textbf{Role activation}: The system activates a small set of relevant social roles including a clinical role concerned with patient safety and standards of care; a patient-advocacy role concerned with quality of life, financial burden, and access barriers; and an institutional role concerned with fair allocation of limited appointment capacity. These roles are not personas. Each is defined by specific responsibilities, sources of evidence, and limits of authority.

\item \textbf{Structured deliberation}: Each role produces a recommendation together with its justification. The clinical role may argue that worsening symptoms justify reassessment. The patient-advocacy role may emphasize that additional visits could create substantial financial hardship. The institutional role may note that urgent appointments should be reserved for patients meeting established escalation criteria. Roles are required to respond to one another's claims so that trade-offs become explicit rather than hidden within a single synthesized answer.

\item \textbf{Coordination}: The objective is not to determine which role ``wins.'' Instead, the system applies an explicit coordination rule. Clinical risk may have priority when evidence suggests potential harm, while patient-burden considerations influence which acceptable pathway is recommended. Institutional considerations constrain recommendations to options that are consistent with existing policies and resource limitations. The system therefore seeks a resolution that respects multiple legitimate perspectives rather than maximizing agreement with any single one.

\item \textbf{Final output}: The resulting recommendation may advise contacting the clinic for reassessment through an existing triage pathway, explain why worsening symptoms warrant review, acknowledge work-related barriers, and identify feasible options that minimize disruption. Rather than presenting multiple disconnected viewpoints, the system provides a coordinated recommendation whose reasoning reflects the interaction of the activated roles.
\end{enumerate}

Pluralistic alignment is therefore not just achieved by generating diverse answers, but by representing socially relevant perspectives, structuring their interaction, and producing decisions through explicit coordination procedures that can be audited, inspected, challenged, and revised.

\section{Conclusion}

Pluralistic alignment has emerged as a critical response to the limitations of single-objective alignment in contemporary AI systems. Existing approaches - Overton, steerable, and distributional pluralism - provide important starting points, but remain limited by an under-specified account of how values are socially organized, contested, and enacted. In this paper, we have argued that social theory offers concrete resources for addressing these limitations.

By drawing on Mead, we reframed Overton pluralism as the modeling of role-structured fields of expectation rather than enumerated viewpoints. Through Habermas, we reconceptualized steerability as the design of deliberative processes that enable structured negotiation among perspectives. Using Bourdieu, we reinterpreted distributional pluralism as a problem of representing fields shaped by power, position, and institutional structure. Finally, we showed how these components must be evaluated and refined through interaction, focusing on trajectories of system behaviour rather than isolated outputs.

These contributions shift pluralistic alignment from an output variation problem to a social coordination problem. This has implications for AI system design, including for role-structured prompting, multi-agent deliberation, population-aware routing, and interaction-level evaluation. Our focus is conceptual, design-oriented, and intended to be operationalizable in existing agentic AI frameworks.

At the same time, socially grounded pluralism must be bounded. Not every expressed preference should be surfaced, weighted, or enacted. Some perspectives fall outside the legitimate action space because they violate safety requirements, rights-based constraints, professional duties, or domain-specific rules. In such cases, the boundary must be made explicit: systems should identify excluded claims, state the limiting constraints, and record whether the issue was resolved by policy, domain authority, or human escalation. This improves legitimacy in high-stakes settings.

The selection of social-theoretical frameworks in this paper is necessarily partial. Mead, Habermas, and Bourdieu provide particularly useful entry points because they articulate complementary accounts of social structure, deliberation, and power. However, they are not exhaustive, nor uniquely suited to this task. The broader claim we advance is that social theory, as a domain, offers a rich set of conceptual and methodological tools to address core challenges in pluralistic alignment. Future work should explore additional traditions, including feminist theory, postcolonial theory, pragmatism, and science and technology studies, to further expand the design space of socially grounded AI.

Further, additional work is needed to implement and empirically evaluate the mechanisms described here in applied domains where plural values are unavoidable, including healthcare, public services, and institutional decision support. Such work will be necessary to assess not only the technical feasibility of socially grounded alignment, but also its effectiveness in producing systems that are more transparent, legitimate, and responsive to the diverse contexts in which they operate.

\bibliography{version1}
\bibliographystyle{icml2026}


\end{document}